%% file: main.tex
\documentclass{article} 
\usepackage{iclr2025_conference,times}

\input{math_commands.tex}

\input{macros}

\usepackage[pagebackref,breaklinks=true,colorlinks,bookmarks=false]{hyperref}

\colorlet{dark-blue}{blue!70!black}
\colorlet{dark-green}{green!80!black}
\colorlet{dark-red}{red!80!black}
\definecolor{cvprblue}{rgb}{0.21,0.49,0.74}
\hypersetup{
    colorlinks=true,%
    citecolor=cvprblue,%
    filecolor=cvprblue,%
    linkcolor=red,%
    urlcolor=magenta
}
\usepackage{hyperref}
\usepackage{url}

\usepackage{graphicx}
\usepackage{amsmath}
\usepackage{amssymb}
\usepackage{booktabs}
\usepackage{caption}

\usepackage{booktabs} 
\usepackage{multirow}
\usepackage{tabularx,verbatim}
\usepackage{xspace}
\usepackage{algorithm}
\usepackage{algorithmic}
\usepackage{setspace}
\usepackage{comment}
\usepackage{bbm}
\usepackage{enumitem}
\usepackage{colortbl}
\usepackage{color}         
\usepackage{listings}

\usepackage{pifont}

\usepackage[capitalize]{cleveref}
\crefname{section}{Sec.}{Secs.}
\Crefname{section}{Section}{Sections}
\Crefname{table}{Table}{Tables}
\crefname{table}{Tab.}{Tabs.}

\usepackage{url}

\iclrfinalcopy

\title{
\raisebox{-0.28\height}{\includegraphics[height=2em]{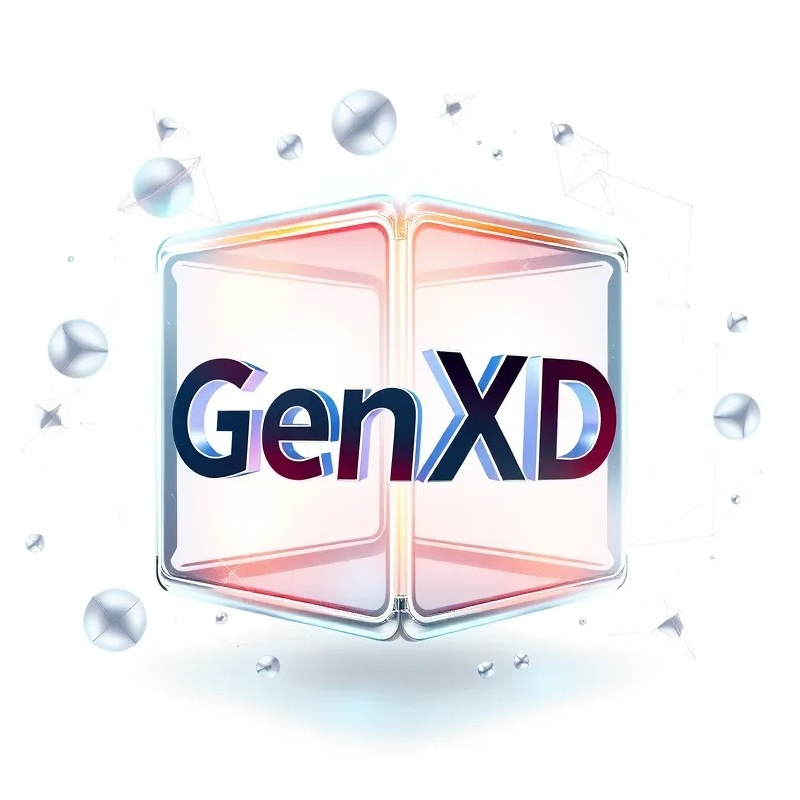}}
\ours: Generating any 3D and 4D Scenes}

\author{\textbf{Yuyang Zhao$^\clubsuit$\thanks{Work was done during internship at Microsoft}~, Chung-Ching Lin$^\spadesuit$, Kevin Lin$^\spadesuit$, Zhiwen Yan$^\clubsuit$, Linjie Li$^\spadesuit$,} \\
\textbf{Zhengyuan Yang$^\spadesuit$, Jianfeng Wang$^\spadesuit$, Gim Hee Lee$^\clubsuit$, Lijuan Wang$^\spadesuit$} \\
$^\clubsuit$ National University of Singapore, $^\spadesuit$ Microsoft Corporation \\
\url{https://gen-x-d.github.io}
}

\begin{document}

\maketitle
\begin{figure}[h]
    \centering
    \includegraphics[width=0.95\linewidth]{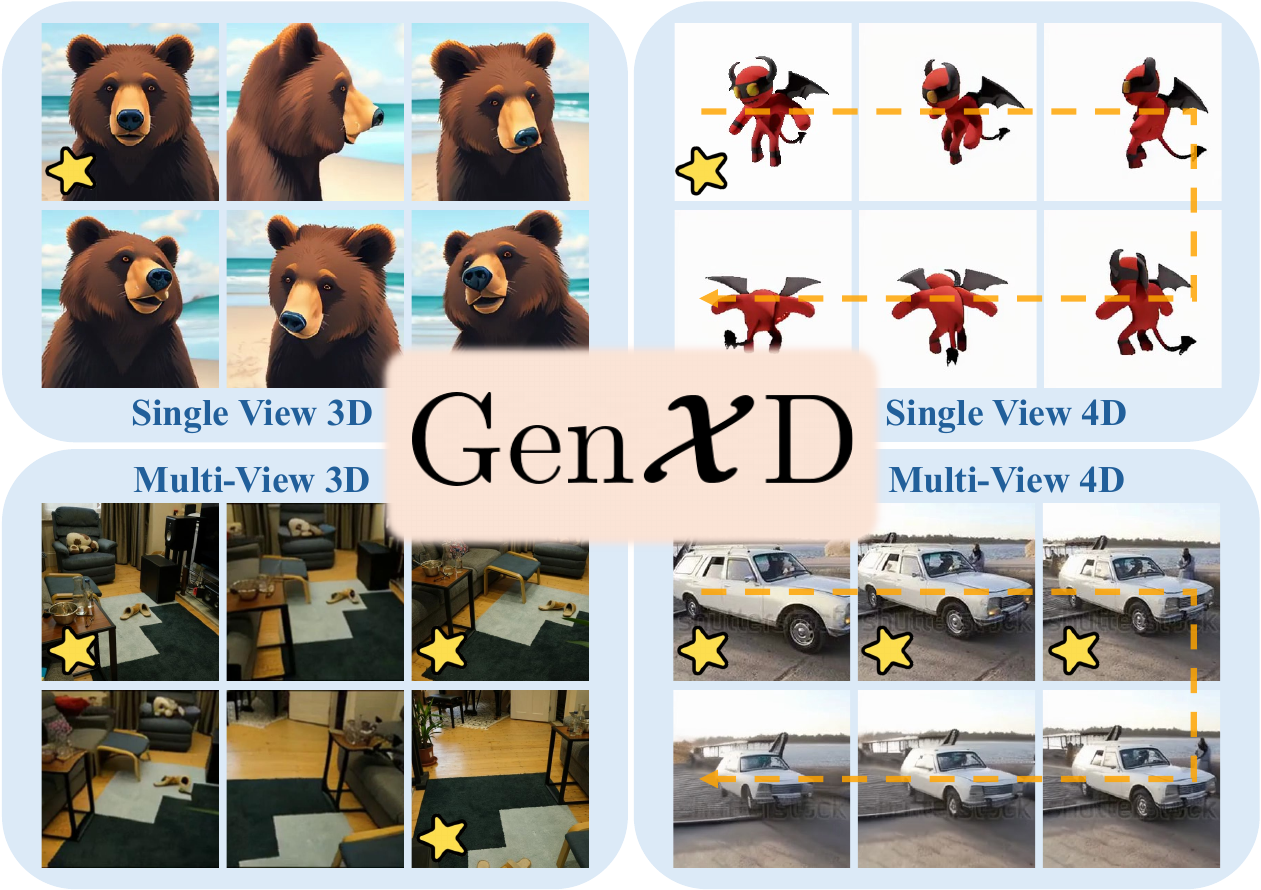}
    \vspace{-.1in}
    \caption{\ours is a unified model for high-quality 3D and 4D generation from any number of condition images. By controlling the motion strength and condition masks, \ours can support various application without any modification. The condition images are shown with \textcolor{yellow}{\textbf{star icon}} and the time dimension is illustrated with \textcolor{celadon}{\textbf{dash line}}.}
    \label{fig:teaser}
\end{figure}

\begin{abstract}

Recent developments in 2D visual generation have been remarkably successful. However, 3D and 4D generation remain challenging in real-world applications due to the lack of large-scale 4D data and effective model design. In this paper, we propose to jointly investigate general 3D and 4D generation by leveraging camera and object movements commonly observed in daily life. Due to the lack of real-world 4D data in the community, we first propose a data curation pipeline to obtain camera poses and object motion strength from videos. Based on this pipeline, we introduce a large-scale real-world 4D scene dataset: \textbf{\data}. By leveraging all the 3D and 4D data, we develop our framework, \textbf{\ours}, which allows us to produce any 3D or 4D scene. We propose multiview-temporal modules, which disentangle camera and object movements, to seamlessly learn from both 3D and 4D data. Additionally, \ours employs masked latent conditions to support a variety of conditioning views. \ours can generate videos that follow the camera trajectory as well as consistent 3D views that can be lifted into 3D representations. We perform extensive evaluations across various real-world and synthetic datasets, demonstrating \ours's effectiveness and versatility compared to previous methods in 3D and 4D generation. The dataset and code will be made publicly available.

\end{abstract}

\section{Introduction}

Generating 2D visual content has achieved remarkable success with diffusion~\citep{LDM,dalle3,sd3,svd} and autoregressive modeling~\citep{var,llamagen,kondratyuk2023videopoet,openmagvit}, which have already been used in real-world applications, benefiting society. In addition to 2D generation, 3D content generation is also of vital importance, with applications in video games, visual effects, and wearable mixed reality devices. However, due to the complexity of 3D modeling and the limitations of 3D data, 3D content generation is still far from satisfactory and is attracting more attention.
In this paper, we focus on the unified generation of 3D and 4D content. Specifically, static 3D content involves only spatial view changes, referred to as \textit{3D generation} in this paper. In contrast, dynamic 3D content includes movable objects within the scene, requiring the modeling of both spatial view and dynamic (temporal) changes, which we term \textit{4D generation}.

Most previous works~\citep{Zero-1-to-3,shi2023mvdream,zhao2023animate124,xie2024sv4d,tang2024lgm,tang2023dreamgaussian} focus on 3D and 4D generation using synthetic object data. Synthetic object data are typically meshes, allowing researchers to render images and other 3D information (\eg, normals and depth) from any viewpoint. However, object generation is more beneficial to specialists than to the general public. In contrast, scene-level generation can help everyone enhance their images and videos with richer content. As a result, recent works~\citep{gao2024cat3d,wu2024reconfusion} have explored general 3D generation (both scene-level and object-level) in a single model, achieving impressive performance.
Nonetheless, these works focus solely on static 3D generation, without addressing dynamics. In this paper, we propose a \textit{unified framework for general 3D and 4D generation}, enabling the generation of images from different viewpoints and timesteps with any number of conditioning images (Fig.~\ref{fig:teaser}).

The first and foremost challenge in 4D generation is the lack of general 4D data. In this work, we propose \textbf{\data}, which contains approximately 30K 4D data samples. 4D data require both multi-view spatial information and temporal dynamics, so we turn to video data to obtain the necessary 4D data. Specifically, we need two key attributes from the video: the camera pose for each frame and the presence of movable objects. To achieve this, we first estimate the possible movable objects in the video using a segmentation model and then estimate the camera pose using keypoints in the static parts of the scene. While successful camera pose estimation ensures multiple views, we also need to ensure that moving objects are present in the video, rather than purely static scenes. To address this, we propose an object motion field that leverages aligned depth to estimate true object movement in the 2D view. Based on the object motion field, we filtered out static scenes, resulting in approximately 30K videos with camera poses.

In addition, we propose a unified framework, \textbf{\ours}, to handle 3D and 4D generation within a single model. While there are similarities between 3D and 4D data in terms of their representation of spatial information, they differ in how they capture temporal information. Therefore, 3D and 4D generation can complement each other through the disentanglement of spatial and temporal information (see Appendix.~\ref{appendix:ablation} for details). To achieve this, we combine both 3D and 4D data during model training.
To disentangle the spatial and temporal information, we introduce multiview-temporal modules in \ours. In each module, we use $\alpha$-fusing to merge spatial and temporal information for 4D data, while removing temporal information for 3D data. Previous works~\cite{xu2024camco,voleti2024sv3d} typically use a fixed number of conditioning images (\eg, the first image). However, single-image conditioning can be more creative, whereas multi-image conditioning offers greater consistency. As a result, we implement masked latent conditioning in our diffusion model. By masking out the noise in the conditioning images, \ours can support any number of input views without modifying the network. With high-quality 4D data and a 4D spatio-temporal generative model, \ours achieves significant performance in both 3D and 4D generation using single or multiple input views.
Our contributions are summarized as follows: 
\begin{itemize} 
\vspace{-.1in} 
\item We design a data curation pipeline for obtaining high-quality 4D data with movable objects from videos and annotate 30,000 videos with camera poses. This large-scale dataset, termed \data, will be made available for public use. 
\item We propose a 3D-4D joint framework, \ours, which supports image-conditioned 3D and 4D generation in various settings (Tab.~\ref{tab:setting}). In \ours, the multiview-temporal layer is introduced to disentangle and fuse multi-view and temporal information. 
\item Using the proposed \data along with other existing 3D and 4D datasets, \ours achieves performance comparable to or better than previous state-of-the-art and baseline methods in single-view 3D object generation, few-view 3D scene reconstruction, single-view 4D generation, and single/multi-view 4D generation. 
\end{itemize}

\input{tables/teaser-table}

\section{Related Work}

\textbf{3D Generation.}
Before the emergence of large-scale 3D data, early works~\citep{RealFusion,DreamFusion} distill the knowledge from 2D diffusion models for text- and image-based 3D generation. Later, with the development of 3D data, 3D generation has been mainly explored in two directions: multi-view priors and feed-forward models. Multi-view priors~\citep{shi2023mvdream,Zero-1-to-3,liu2023syncdreamer,long2024wonder3d,gao2024cat3d} generate multi-view images and other features (\eg, normal maps and depths) based on camera embeddings, and then train a 3D representation using the generated samples or distill from generative priors. Feed-forward models~\citep{hong2023lrm,liu2024meshformer,one-2-3-45,tang2024lgm,xray} directly predict NeRF~\citep{hong2023lrm}, 3D Gaussians~\citep{tang2024lgm,tochilkin2024triposr}, or meshes~\citep{one-2-3-45,liu2024meshformer} from single or multi-view images. Compared to multi-view priors, feed-forward models are more efficient but produce lower quality results. In this paper, we follow the paradigm of multi-view priors.

\textbf{4D Generation.}
Similar to 3D generation, early 4D generation works~\citep{zhao2023animate124,mav3d,ling2024align,ren2023dreamgaussian4d} distill 2D video generation models into 4D representation. Due to the variability and complexity of multi-view videos, these methods typically require long optimization time. Later, by leveraging animated 3D mesh data~\citep{objaversexl}, researchers render multi-view videos and use them to train 4D diffusion models~\citep{liang2024diffusion4d} and feed-forward models~\citep{ren2024l4gm}. Although these models achieve good multi-view and video quality, they only focus on object-centric synthetic scenarios rather than entire scenes. This limitation is due to the lack of scene-level 4D data and the requirement for both multi-view static and dynamic information in these models.
In this paper, we introduce a large-scale dataset of scene-level 4D data and address the challenge of general 4D generation.

\textbf{Camera-controlled Video Generation.}
In the real world, videos contain not only object motion but also camera movement. Therefore, controlling camera movement in video generation has also garnered attention in the community~\citep{wang2024motionctrl,xu2024camco,he2024cameractrl}. MotionCtrl~\cite{wang2024motionctrl} and CameraCtrl~\cite{he2024cameractrl} introduce a branch to encode camera information from multi-view 3D data and integrate it into a frozen video generation model. However, due to the limitations of this integration approach, these methods cannot generate videos that align well with the camera pose.
CamCo~\cite{xu2024camco} annotates some 4D data similar to ours and fine-tunes the entire video generation model using this data. However, due to limitations in camera pose quality and diversity, CamCo struggles to handle large camera movements.

\begin{figure}[t]
    \centering
    \includegraphics[width=.99\linewidth]{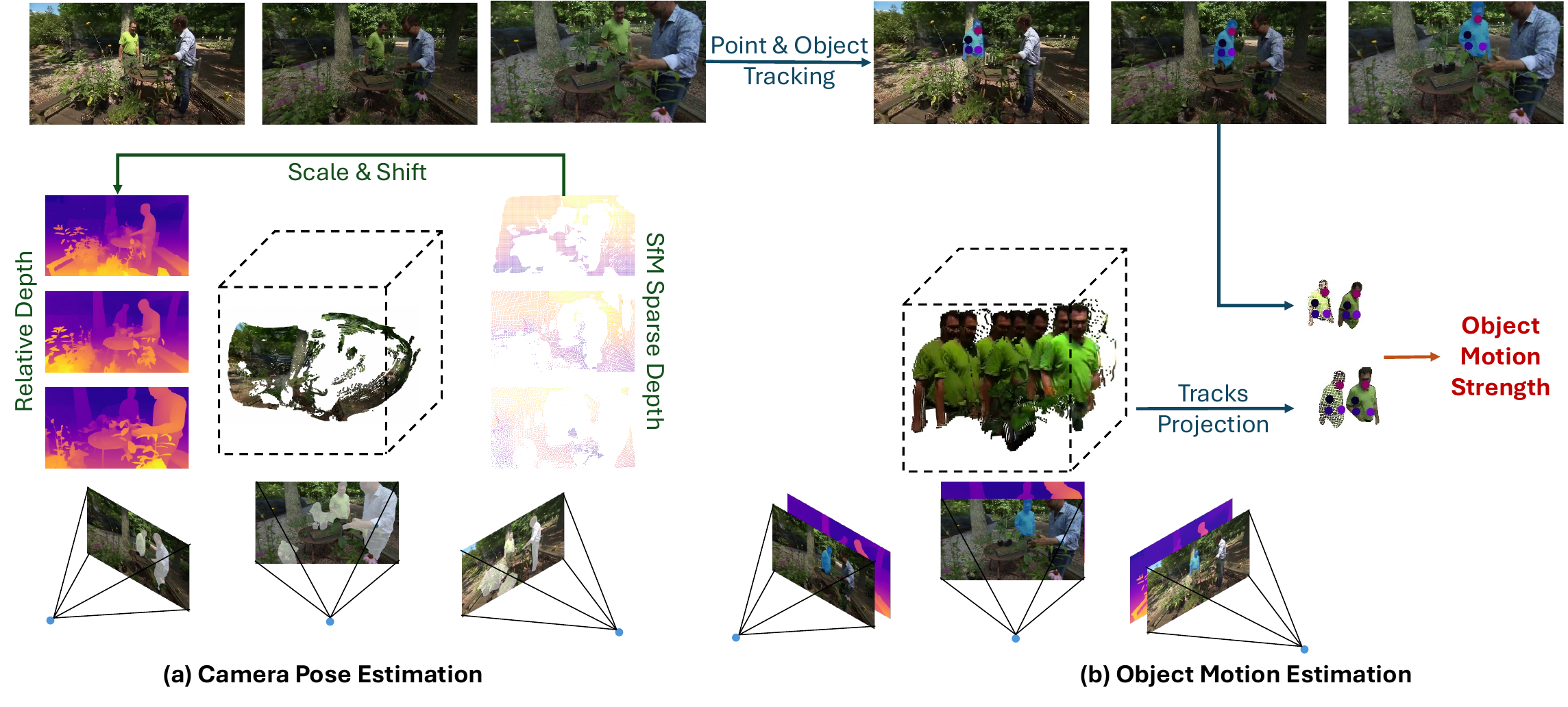}
    \vspace{-.1in}
    \caption{\textbf{The pipeline for \data data curation}, including (a) camera pose estimation and (b) object motion estimation. We first leverage mask-based SfM (masks are overlayed to images in (a) for visualization) to estimate camera pose and reconstruct 3D point clouds of static parts. Then relative depth is aligned with the sparse depth and project the tracking keypoints to consecutive frame for object motion estimation.}
    \label{fig:motion-estimation}
    \vspace{-.1in}
\end{figure}

\section{CamVid-30K}
\label{sec:camvid}

The lack of large-scale 4D scene data limits the development of dynamic 3D tasks, including but not limited to 4D generation, dynamic camera pose estimation, and controllable video generation. To address this, we introduce a high-quality 4D dataset in this paper. First, we estimate the camera poses using a Structure-from-Motion (SfM) based method, then filter out data without object movement using the proposed motion strength. The pipeline is illustrated in Fig.~\ref{fig:motion-estimation}

\subsection{Camera Pose Estimation}

The camera pose estimation is based on SfM, which reconstructs 3D structure from their projections in a series of images. SfM involves three main steps: (1) feature detection and extraction, (2) feature matching and geometric verification, and (3) 3D reconstruction and camera pose estimation. In the second step, the matched features must be on the \textbf{static} part of the scene. Otherwise, object movement will be interpreted as camera movement during feature matching, which can impair the accuracy of camera pose estimations.

To address this, Particle-SfM~\cite{zhao2022particlesfm} separates moving objects from the static background using a motion segmentation module, and then performs SfM on the static part to estimate camera poses. However, it is extremely difficult to accurately detect moving pixels when the camera itself is moving, and we empirically observe that the motion segmentation module in \citet{zhao2022particlesfm} lacks sufficient generalization, leading to false negatives and incorrect camera poses. To obtain accurate camera poses, it is essential to segment \textbf{all} moving pixels. In this case, a \textit{false positive} error is more acceptable than a false negative.
To achieve this, we use an instance segmentation model~\cite{mask2former} to greedily segment all pixels that might be moving. The instance segmentation model is far more generalizable than the motion segmentation module in \citet{zhao2022particlesfm}, particularly on training categories. After segmenting the potentially moving pixels, we estimate the camera pose with Particle-SfM~\cite{zhao2022particlesfm} to obtain camera information and sparse point clouds (Fig.~\ref{fig:motion-estimation}(a)).

\subsection{Object Motion Estimation}
\label{sec:obj-motion}

\textbf{Unravel Camera and Object Motion.}
While instance segmentation can accurately separate objects from backgrounds, it cannot determine whether the object itself is moving, and static objects negatively impact motion learning. Thus, we introduce motion strength to identify true object motion and filter out videos with only static objects.

Since camera movement and object motion are both present in videos, 2D-based motion estimation methods (\eg, optical flow) cannot accurately represent true object motion. There are two ways to capture true object motion: by measuring motion in 3D space or by projecting motion in videos to the same camera. Both approaches require depth maps aligned with the camera pose scale. The sparse depth map can be obtained by projecting 3D point clouds $P_{\text{world}}$ onto the camera view:
\begin{equation}
    \label{eq:projection}
    P_{\text{camera}} = R \cdot P_{\text{world}} + t, \qquad (u,v,1)^T = K \cdot (X_c/Z_c, Y_c/Z_c, 1)^T,
\end{equation}
where $P_{\text{camera}} = (X_c, Y_c, Z_c)$ denotes the coordinates of point cloud $c$ in the camera space. 
$R$ and $t$ denote the rotation and translation to transform from world space to camera space. $K$ is the camera intrinsics. With the projection equation, the depth value $d_{\text{SfM}}$ on the image pixel $(u,v)$ can be obtained by $d_{\text{SfM}}(u,v) = Z_c$.

As shown in Fig.~\ref{fig:motion-estimation}(a), since only features in the static parts are matched during 3D reconstruction, we can only obtain sparse point clouds for the static regions. However, depth information in the dynamic parts is crucial for estimating motion. To address this, we leverage a pre-trained relative monocular depth estimation model~\cite{depthanythingv2} to predict the relative depth of each frame $d_{\text{rel}} \in [0,1]$. We then apply a scale factor $\alpha$ and a shift $\beta$ to align it with the SfM sparse depth:
\begin{equation}
\begin{aligned}
    \label{eq:depth-align}
    \alpha = \text{median}(d_{\text{SfM}}) / \text{median}(d_{\text{rel}}), &\qquad \beta = \text{median}(d_{\text{SfM}} - \alpha \cdot d_{\text{rel}}), \\
    d_{\text{aligned}} &= \alpha * d_{\text{rel}} + \beta,
\end{aligned}
\end{equation}
where $\text{median}(\cdot)$ denotes the median value, and $d_{\text{aligned}}$ is the dense depth map aligned with SfM depth scale. 

\begin{figure}[t]
    \centering
    \includegraphics[width=0.9\linewidth]{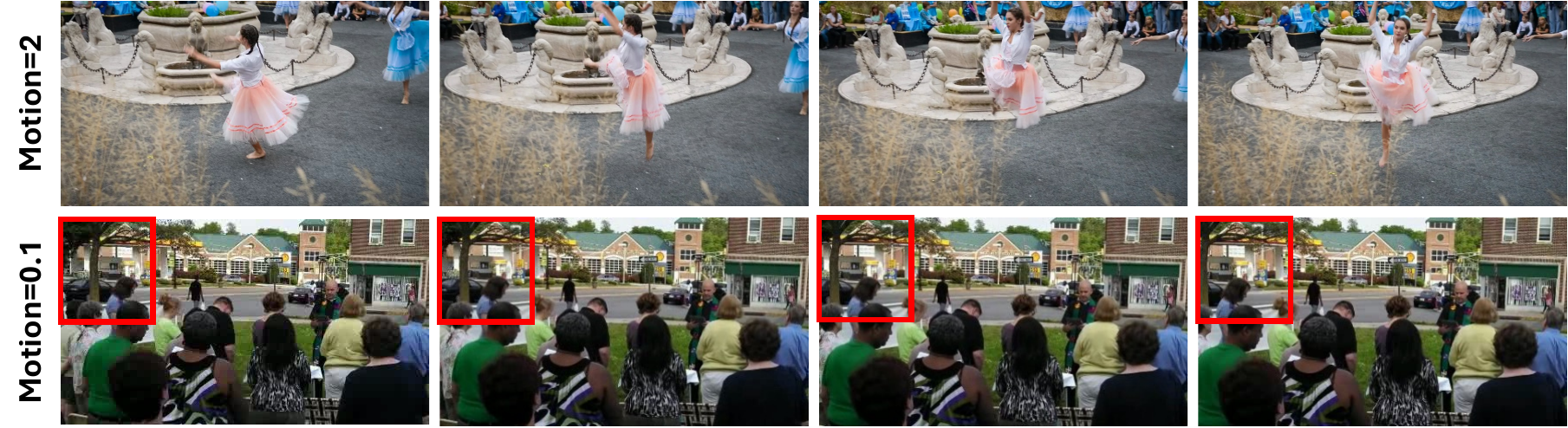}
    \vspace{-.1in}
    \caption{\textbf{Examples for object motion estimation.} 
    The motion strength is multiplied by 100. In the first example, the girl is dancing, together with the camera moving. In the second example, the camera is zooming in (\textcolor{red}{red} rectangle for better illustration) but the object is static. In this case, the motion strength is much smaller.}
    \label{fig:motion-example}
    \vspace{-.1in}
\end{figure}

\textbf{Object Motion Field.}
With the aligned depth $d_{\text{align}}$, we can project dynamic objects in a frame into 3D space, providing a straightforward way to measure object motion. As shown in Fig.~\ref{fig:motion-estimation}(b), if the object (\eg, the man in the green shirt) is moving, there will be displacement in the projected 3D point clouds. However, since SfM operates up to a scale, measuring motion directly in 3D space can lead to magnitude issues. Therefore, we project the dynamic objects into adjacent views and estimate the object motion field.

Specifically, we first need to find matching points in the 2D video. Instead of using dense representations like optical flow, we sample keypoints for each object instance and use video object segmentation~\cite{vos} and keypoint tracking~\cite{pointtracking} in 2D videos to establish matching relationships. Each keypoint is then projected into adjacent frames. The keypoint $(u_i,v_i)^T$ in the $i$-th frame is first back-projected into world space to obtain the 3D keypoint $kp_i$:
\begin{equation}
    \label{eq:back-projection}
    kp_i = Z_i \cdot K^{-1} \cdot (u_i, v_i, 1)^T,
\end{equation}
where $Z_i=d_{\text{aligned}}(u_i,v_i)$ is the depth value in the aligned dense depth map.
Then the 3D keypoint is projected to $j$-th frame with the projection equation (Eq.~\ref{eq:projection}) to obtain the 2D projected keypoint $(u_{ij},v_{ij})^T$. Similar to optical flow, we represent the displacement of each 2D keypoint on the second camera view as object motion field:
\begin{equation}
    \label{eq:motion-field}
    (\Delta u_{ij},\Delta v_{ij})^T = ((u_j - u_{ij}) / W, (v_j - v_{ij}) / H)^T,
\end{equation}
where $H$ and $W$ denotes image height and width.

With the motion field for each object, we can estimate the global movement of an object by averaging the absolute magnitude of the motion field. For each video, the motion strength is represented by the maximum movement value among all the objects. As shown in Fig.~\ref{fig:motion-example}, when the camera is moving while the object remains static (second example), the motion strength is significantly smaller compared to videos with object motion. Using motion strength, we further filter out data that lacks obvious object movement. The motion strength value also serves as a good indicator of the scale of object movement, which is used in the temporal layer to enable better motion control (Sec.~\ref{sec:generativemodel}).

\begin{figure}[t]
    \centering
    \includegraphics[width=.99\linewidth]{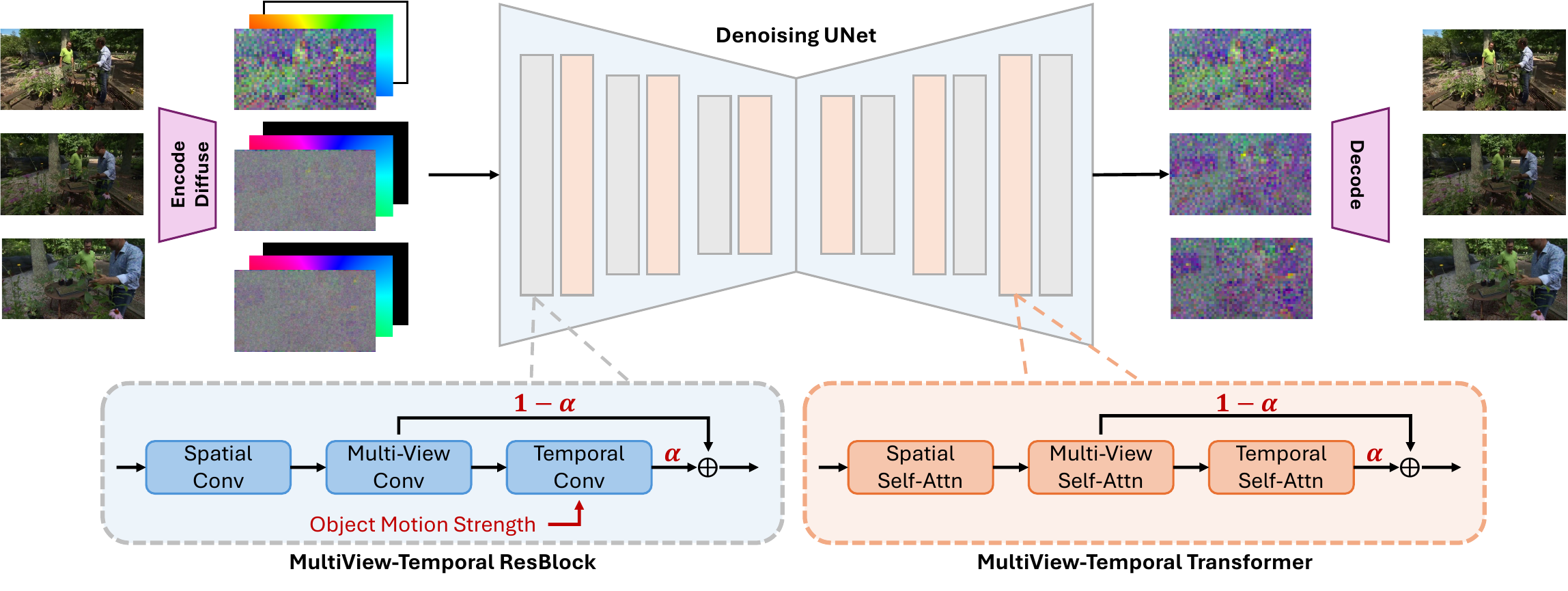}
    \vspace{-.1in}
    \caption{\textbf{The framework of \ours.} 
    We leverage mask latent conditioned diffusion model to generate 3D and 4D samples with both camera (colorful map) and image (binary map) conditions. In addition, multiview-temporal modules together with $\alpha$-fusing are proposed to effectively disentangle and fuse multiview and temporal information.}
    \label{fig:diffusion}
    \vspace{-.15in}
\end{figure}

\section{\ours}
\subsection{Generative Model}
\label{sec:generativemodel}

Since most scene-level 3D and 4D data are captured via videos, these data lack explicit representations (\eg, meshes). Therefore, we adopt an approach that generates images aligned with spatial camera poses and temporal timesteps. Specifically, we incorporate the latent diffusion model~\cite{LDM} into our framework, introducing additional multiview-temporal layers, including multiview-temporal ResBlocks and multiview-temporal transformers, to disentangle and fuse 3D and temporal information.

\textbf{Mask Latent Conditioned Diffusion Model.}
Latent diffusion model (LDM)~\cite{LDM} is leveraged in \ours to generate images of different camera viewpoint and time together. LDM first encode an image/video into a latent code $z$ with VAE~\cite{vae} and diffuse the latent with gaussian noise $\epsilon$ to obtain $z_t$. Then a denoising model $\epsilon_\theta(\cdot)$ is leveraged to estimate the noise and reverse the diffusion process with conditions:
\begin{equation}
\label{eq:ldm}
L_{\text{LDM}}:=\mathbb{E}_{\mathcal{E}(x), \epsilon \sim \mathcal{N}(0,1), t}\left[\left\|\epsilon-\epsilon_\theta\left(z_t, t,c\right)\right\|_2^2\right],
\end{equation}
where $c$ is the condition used for controllable generation, which is commonly text or image. 

\ours generates multi-view images and videos with camera pose and reference image, and thus it requires both camera and image conditions. Camera conditions are independent for each image, either conditioned or targeted. Therefore, it is easy to append it to each latent. Here, we opt for Pl\"ucker ray~\cite{pluckerray} as camera condition:
\begin{equation}
\mathbf{r}=\langle\mathbf{d}, \mathbf{o}\times \mathbf{d}\rangle \in \mathbb{R}^6
\end{equation}
where $\mathbf{o}\in\mathbb{R}^3$ and $\mathbf{d}\in\mathbb{R}^3$ denote the camera center and the ray direction from camera center to each image pixel, respectively.
Therefore, Pl\"ucker ray is a dense embedding encoding not only the pixel information, but also the camera pose and intrinsic information, which is better than global camera representation. 

The reference image condition is more complex. \ours aims to conduct 3D and 4D generation with both single and multiple input views. The single view generation has less requirement while the multi-view generation has more consistent results. Therefore, combining single and multi-view generation will lead to better real-world application. However, previous works~\cite{svd,voleti2024sv3d,Zero-1-to-3} condition images by concatenating condition latent to the target latents and by incorporating CLIP image embedding~\cite{CLIP} via cross attention. The concatenation way requires to change the channel of the model, which is unable to process arbitrary input views. The CLIP embedding can support multiple conditions. However, both ways cannot model the positional information of multiple conditions and model the information among the input views. In view of the limitations, we leverage the mask latent conditioning~\cite{gao2024cat3d,videointerpolation} for image conditions. As shown in Fig.~\ref{fig:diffusion}, after encoding with VAE encoder, the forward diffusion process is applied to the target frames (2nd and 3rd frame), leaving the condition latent (1st frame) as usual. Then the noise on the two frames are estimated by denoising model and removed by the backward process. 

The mask latent conditioning has three main benefits.
First, model can support any input views without modification on the parameters. 
Second, for sequence generation (multi-view images or video), we do not need to constraint the position of the condition frame since the condition frame keeps its position in the sequence. In contrast, many works~\cite{svd,xu2024camco,chen2023videocrafter1} requires the condition image at a fixed position in the sequence (commonly the first frame). 
Third, without the conditioning embedding from additional models~\cite{CLIP}, the cross attention layers used to integrate conditioning embedding can be removed, which will greatly reduce the number of model parameters. To this end, we leverage mask latent conditioning approach for \ours.

\textbf{MultiView-Temporal Modules.}
As \ours aims to generate both 3D and 4D samples within a single model, we need to disentangle the multi-view information from the temporal information. We model these two types of information in separate layers: the multi-view layer and the temporal layer. For 3D generation, no temporal information is considered, while both multi-view and temporal information are required for 4D generation. Therefore, as illustrated in Fig.~\ref{fig:diffusion}, we propose an \textbf{$\alpha$-fusing} strategy for 4D generation. Specifically, we introduce a learnable fusing weight, $\alpha$, for 4D generation, with $\alpha$ set to 0 for 3D generation. Using the $\alpha$-fusing strategy, \ours can preserve the multi-view information in the multi-view layer for 3D data while learning the temporal information from 4D data.

$\alpha$-fusing can effectively disentangle the multi-view and temporal information. However, the motion is less controllable without any cues. Video generation models~\cite{zhou2022magicvideo,svd} use FPS or motion id to control the magnitude of motion without considering the camera movement.
Thanks to the motion strength in \data, we can effectively represent the object motion.
Since the motion strength is a constant, we combine it with the diffusion timestep and add it to the temporal resblock layer, as illustrated in MultiView-Temporal ResBlock of Fig.~\ref{fig:diffusion}. With the multiview-temporal modules, \ours can effectively conduct both 3D and 4D generation.

\subsection{Generation with 3D Representation}

\ours can generate images with different viewpoints and timesteps using one or several condition images. However, to render arbitrary 3D-consistent views, we need to lift the generated samples into a 3D representation. Previous works~\cite{wu2024reconfusion,shi2023mvdream,zhao2023animate124} commonly optimize 3D representations by distilling knowledge from generative models. Since \ours can generate high-quality and consistent results, we directly use the generated images to optimize the 3D representation. Specifically, we utilize 3D Gaussian Splatting (3D-GS)\cite{3dgs} and Zip-NeRF\cite{barron2023zip} for 3D generation, and 4D Gaussian Splatting (4D-GS)\cite{4dgs} for 4D generation. More details can be found in Appendix.~\ref{appendix:details}.

\section{Experiment}

\subsection{Experimental Setup}
\label{sec:experiment-setup}

\textbf{Datasets.}
\ours is trained with the combination of 3D and 4D datasets. For 3D datasets, we leverage five datasets with camera pose annotation: Objaverse~\cite{objaverse}, MVImageNet~\cite{mvimagenet}, Co3D~\cite{co3d}, Re10K~\cite{re10k} and ACID~\cite{acid}. Objaverse is a synthetic dataset with meshes, and we render the 80K subset~\cite{tang2024lgm} from 12 views following~\cite{Zero-1-to-3}. 
MVImageNet and Co3D are video data recording objects in 239 and 50 categories, respectively. Re10K and Acid are video data that record real-world indoor and outdoor scenarios. For 4D datasets, we leverage the synthetic data Objaverse-XL-Animation~\cite{objaversexl,liang2024diffusion4d} and our \data.
For the Objaverse-XL-Animation, we use the subset filtered by \citet{liang2024diffusion4d}, and re-render the depth and images by adding noise to the oribit camera trajectory. With the ground truth depth, we estimate the object motion strength as in Sec.~\ref{sec:obj-motion}, and then filter out data without obvious object motion. Finally, we get 44K synthetic data from Objaverse-XL-Animation and 30K real-world data from \data.

\textbf{Implementation Details.}
\ours is partially initialized from Stable Video Diffusion (SVD) pre-trained model~\cite{svd} for fast convergence. 
Specifically, both the multi-view layer (multi-view convolution and multi-view self-attention) and temporal layer (temporal convolution and temporal self-attention) are initialized from the temporal layer in SVD, and the cross-attention layers in SVD are removed.
\ours is trained in three stages. We first train the UNet only with 3D data for 500K iteration and then fine-tune it with both 3D and 4D data for 500K iterations in single view mode. Finally, \ours is trained with both single view and multi-view mode with all the data for 500K iteration.
The model is trained on 32 A100 GPUs with batch size 128 and resolution 256$\times$256. AdamW~\cite{loshchilov2019adamw} optimizer with learning rate $5\times 10^{-4}$ is adopted. In the first stage, data are center cropped to square. In the final stage, we make the images square by either center crop or padding, leading to \ours working well on different image ratio.

\begin{figure}[t]
    \centering
    \includegraphics[width=0.9\linewidth]{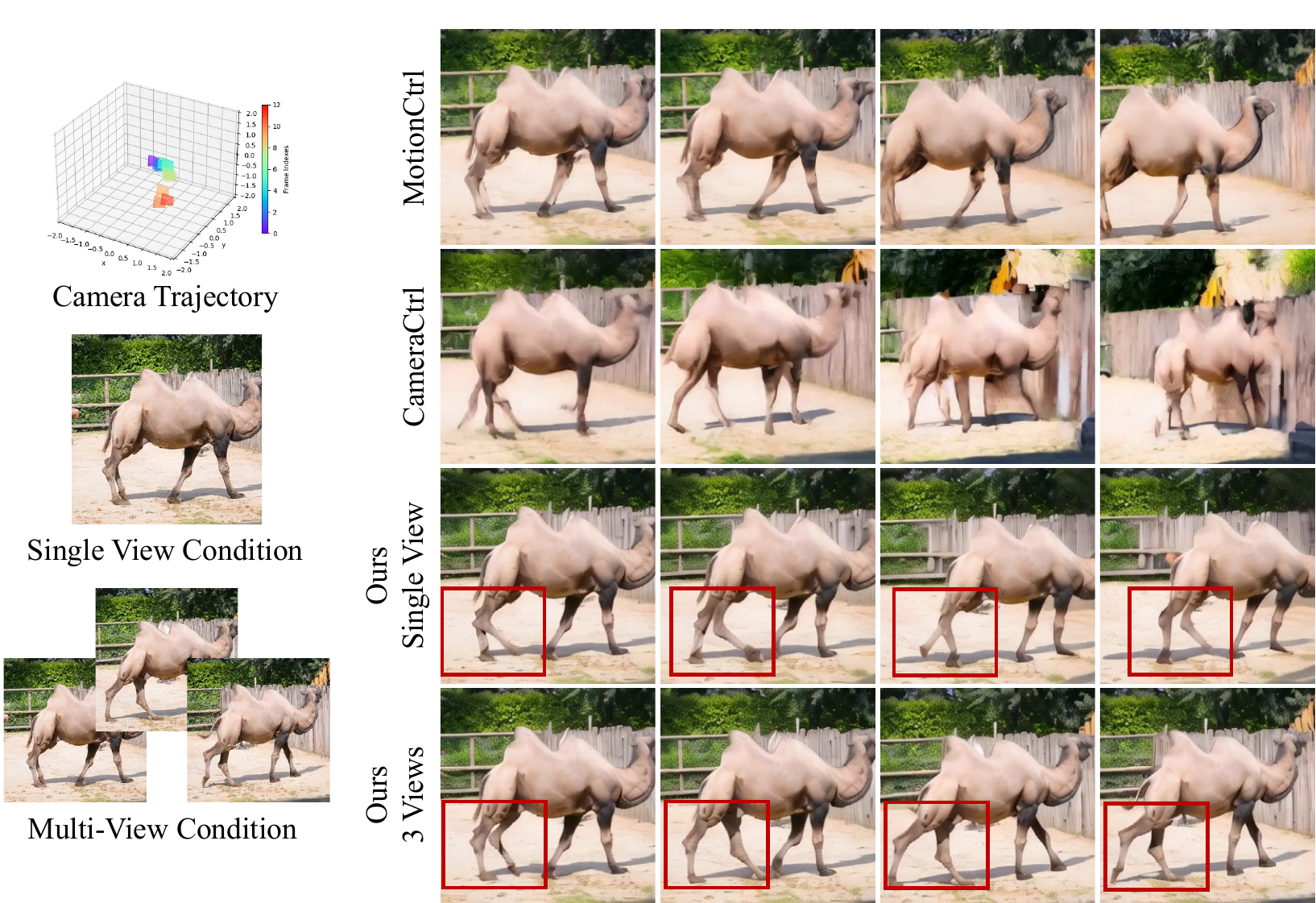}
    \vspace{-.1in}
    \caption{\textbf{Qualitative comparison with camera conditioned video generation methods.} \ours can generate video well-aligned with camera trajectory and containing realistic object motion. (Please refer to supplementary video for better illustration.)}
    \label{fig:4dscene}
    \vspace{-.15in}
\end{figure}

\input{tables/4d}

\subsection{4D Generation}

\textbf{4D Scene Generation.}
In this setting, videos with both object and camera movement are required for evaluation. Therefore, we introduce the Cam-DAVIS benchmark for 4D evaluation. Specifically, we use our annotation pipeline proposed in Sec.~\ref{sec:camvid} to get the camera poses for videos in DAVIS~\cite{davis} dataset. Then we filter the data and get 20 videos with both accurate camera poses and obvious object movement.
The camera trajectories of Cam-DAVIS are out-of-distribution from the training data and thus are good evaluation for the robustness to the camera movement.

We compare \ours with the open-sourced camera conditioned video generation methods, MotionCtrl~\cite{wang2024motionctrl} and CameraCtrl~\cite{he2024cameractrl}, on FID~\cite{fid} and FVD~\cite{fvd} evaluation metrics. We use Stable Video Diffusion~\cite{svd} as the base model for both previous methods and generate the videos with the camera trajectory and the first frame conditions.
As shown in Tab.~\ref{tab:4d-scene}, using first view as condition, \ours outperforms CameraCtrl and MotionCtrl in terms of both metrics significantly. In addition, with 3 views as conditions (first, middle and last frames), \ours outperforms previous works by a large margin. Such results demonstrate the strong generalization ability of \ours on 4D generation.
In Fig.~\ref{fig:4dscene}, we compare the qualitative results of the three methods. In this example, MotionCtrl cannot generate obvious object motion and the video generated by CameraCtrl is neither 3D nor temporal consistent. Instead, our single view conditioned model can generate smooth and consistent 4D video. With 3 conditioning views, \ours can generate quite realistic results.

\textbf{4D Object Generation.}
We evaluate the performance on 4D object generation following \citet{zhao2023animate124}. Since \ours only leverages image condition instead of image-text condition as Animate124~\cite{zhao2023animate124}, we compare the optimization time and CLIP image similarity in Tab.~\ref{tab:4d-object}.
Instead of optimizing dynamic NeRF with score distillation sampling (SDS)~\cite{DreamFusion}, \ours directly generates 4D videos of the orbit camera trajectory and uses such videos to optimize the 4D-GS~\cite{4dgs}. 
This results in our method being 100$\times$ faster than Animate124.
In addition, the semantic drift problem mentioned in \citet{zhao2023animate124} is well addressed in \ours since we use the condition image for 4D generation.
The results on 4D scene and object generation demonstrate the superiority of \ours in generating 3D and temporal consistent 4D videos.

\begin{figure}[t]
    \centering
    \includegraphics[width=0.85\linewidth]{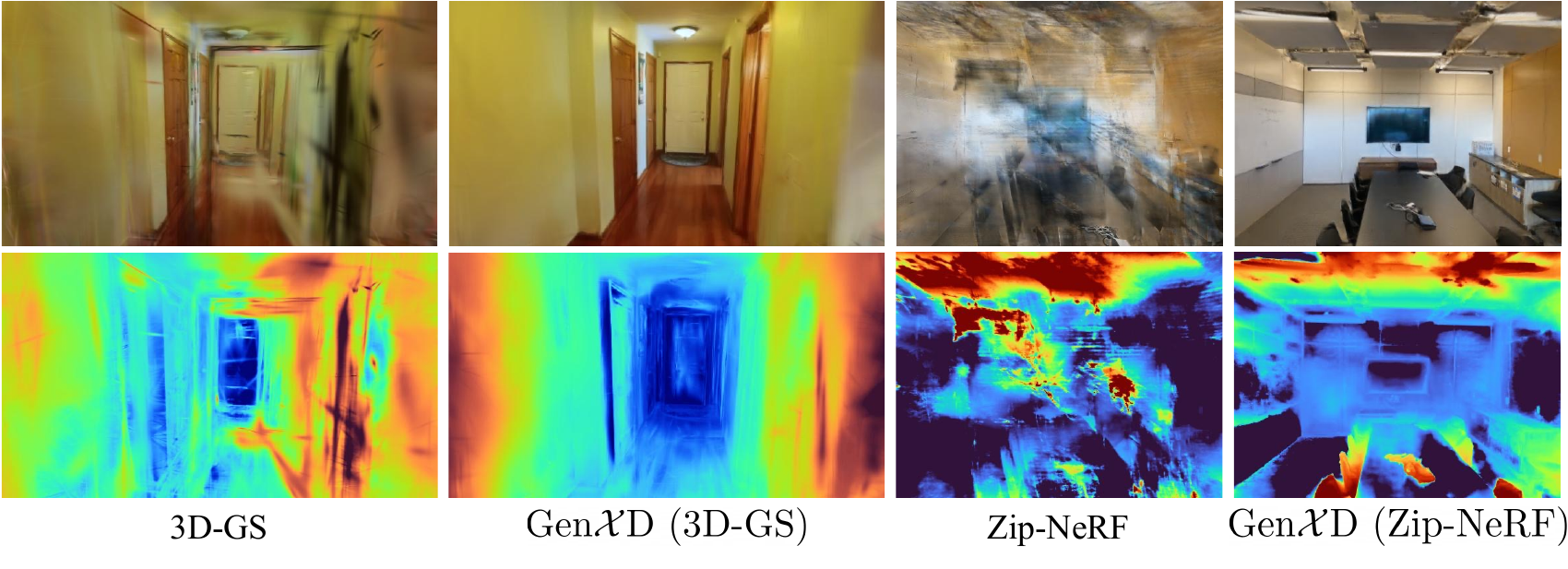}
    \vspace{-.2in}
    \caption{\textbf{Qualitative comparison of few-view 3D reconstruction.}}
    \label{fig:few-view-3d}
    \vspace{-.1in}
\end{figure}

\input{tables/few-view3d}

\subsection{3D Generation}

\textbf{Few View 3D Generation.}
For few-view 3D reconstruction setting, we evaluate \ours on both in-distribution (Re10K~\cite{re10k}) and out-of-distribution (LLFF~\cite{llff}) datasets. We select 10 scenes from Re10K and all the 8 scenes in LLFF, and 3 views in each scene are used for training. The performance is evaluated with PSNR, SSIM and LPIPS metrics on the rendered test views. 
As a generative model, \ours can generate additional views from sparse input views and improve the performance of any reconstruction method. In this experiment, we leverage two baseline methods: Zip-NeRF~\cite{barron2023zip} and 3D-GS~\cite{3dgs}.
The two baselines are methods for many-view reconstruction, and thus we adjust the hyperparameter for better few-view reconstruction (more details in Appendix.~\ref{appendix:details}). 
As shown in Tab.~\ref{tab:few-view3d}, both Zip-NeRF and 3D-GS can be improved with the generated images from \ours, and the improvement is more significant with the Zip-NeRF baseline. Specifically, the PSNR on Re10K (in-distribution) and LLFF (out-of-distribution) are increased by 4.82 and 5.13.
The qualitative comparison is illustrated in Fig.~\ref{fig:few-view-3d}. With the generated views, the floaters and blurs are reduced in the reconstructed scene.
We also evaluate the performance on single view generation setting in Appendix.~\ref{appendix:single-view}.

\subsection{Ablation Study}

In this section, we conduct the ablation study of multiview-temporal modules. The ablation study is evaluated on the quality of the generated diffusion samples on few view 3D and single view 4D generation settings (Tab.~\ref{tab:ablation}).
More ablation studies are conducted in Appendix.~\ref{appendix:ablation}

\begin{figure}
    \centering
    \includegraphics[width=0.9\linewidth]{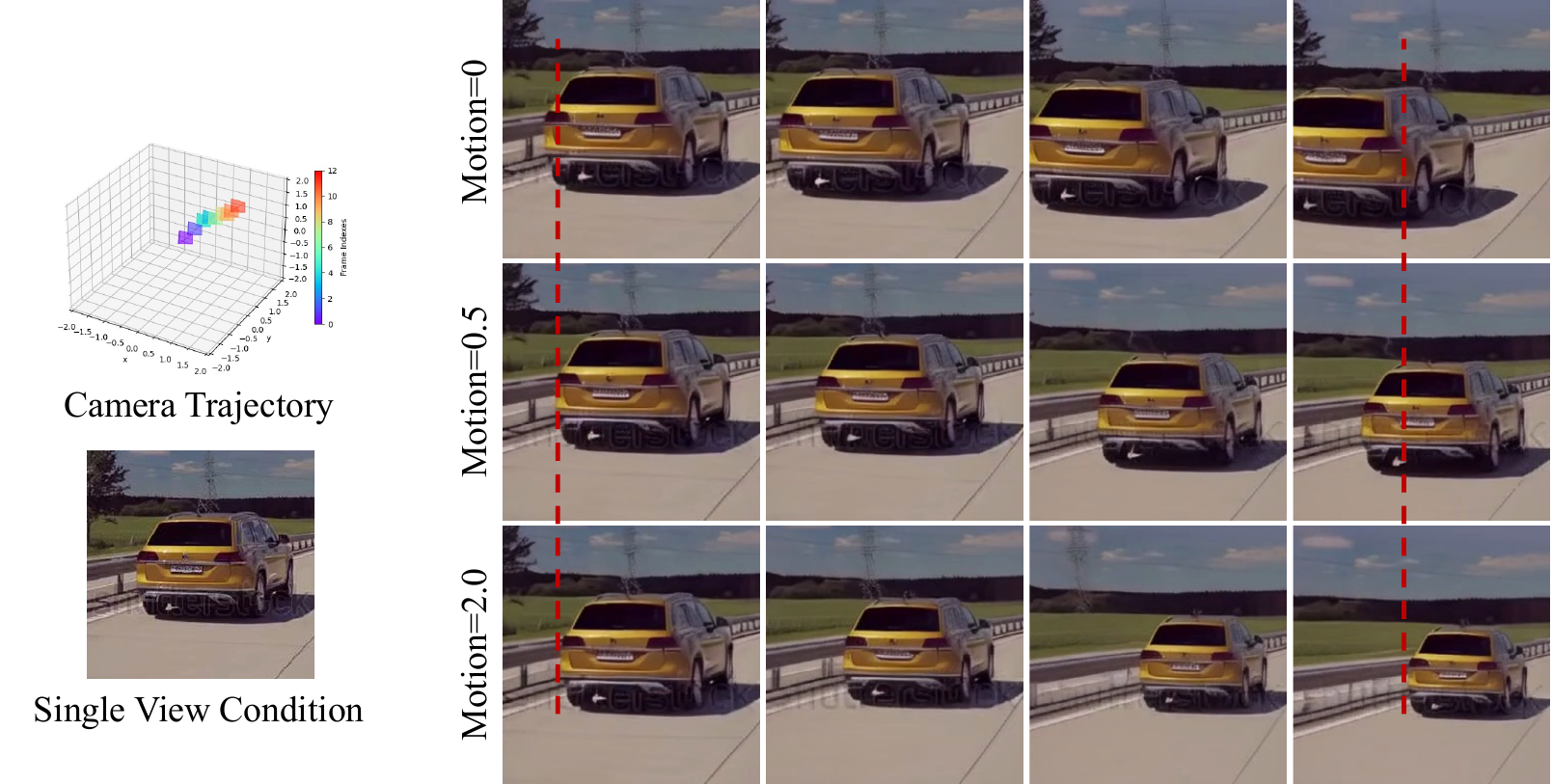}
    \vspace{-.1in}
    \caption{\textbf{Qualitative evaluation on the influence of motion strength.} (Please refer to supplementary video for better illustration.)}
    \vspace{-.1in}
    \label{fig:motion-ablation}
\end{figure}

\input{tables/ablation}

\textbf{Motion Disentangle ($\alpha$-fusing).}
The camera movement and object motion are entangled in 4D data. To enable high quality generation in both 3D and 4D, \ours introduces multiview-temporal modules (Sec.~\ref{sec:generativemodel}) to learn the multi-view and temporal information separately, and then fuse them together with $\alpha$-fusing. For 3D generation, the $\alpha$ is set to 0 to bypass the temporal module while the $\alpha$ is learned during training for 4D generation. 
{Removing the $\alpha$-fusing will result in all 3D and 4D data passing through temporal modules, which will result in the model being unable to disentangle object motion from camera movement. The failure of disentanglement will adversely affect both 3D and 4D generation (Tab.~\ref{tab:ablation}).}

\textbf{Effectiveness of Motion Strength.}
{The motion strength can be used to effectively control the magnitude of the object's movement.} As shown in the second to last row of Fig.~\ref{fig:motion-ablation}, increasing the motion strength can increase the speed of the car.
{As a result of these observations, we can conclude that it is important to learn object motion and that the object motion field and motion strength in our data curation pipeline can accurately represent true object motion.}

\section{Conclusion}
In this paper, we investigate the general 3D and 4D generation with diffusion models. 
To enhance the learning of 4D generation, we first propose a data curation pipeline to annotate camera and object movement in the videos. Equipped with the pipeline, the largest real-world 4D scene dataset, \data, is introduced in this paper. Furthermore, leveraging the large-scale datasets, we propose \ours to handle general 3D and 4D generation. \ours utilize multiview-temporal modules to disentangle camera and object movement and is able to support any number of input condition views by mask latent conditioning. \ours can handle versatile applications and can achieve comparable or better performance in all settings with one model.

\section*{Ethics Statement}

In this paper, we introduce a 4D dataset, \data, and a generative model for general 3D and 4D generation. \data is curated from existing public video datasets~\cite{nan2024openvid,webvid,vipseg}, and we additionally estimate the camera poses and object motion. \data adheres to the licenses and agreements of the original video datasets, and it does not raise any new ethical concerns.
While we ensure compliance with the licenses of the curated datasets and advocate for responsible use, the ability to generate realistic images and videos from various viewpoints raises risks related to misinformation and privacy violations. We encourage the development of tools for detecting misuse, along with responsible dissemination of the dataset and model, to balance innovation with ethical considerations.

\section*{Reproducibility Statement}
The experiments in our paper mainly include the training of \ours and generation with 3D representation in different settings. In Sec.~\ref{sec:experiment-setup}, we describe the 3D and 4D datasets used to train the diffusion model, together with the training configurations.
In Appendix.~\ref{appendix:details}, we introduce the backbone models and the implementation details for generation with 3D representation in each setting.
Our curated 4D dataset, \data, and \ours model will be made publicly available.

\section*{Acknowledgements}
We would like to thank Dejia Xu and Yuyang Yin for their valuable discussions on the 4D data.

\bibliography{reference}
\bibliographystyle{iclr2025_conference}

\newpage
\appendix

\section*{Appendix}

\section{More Qualitative Results}

\begin{figure}[h]
    \centering
    \includegraphics[width=0.9\linewidth]{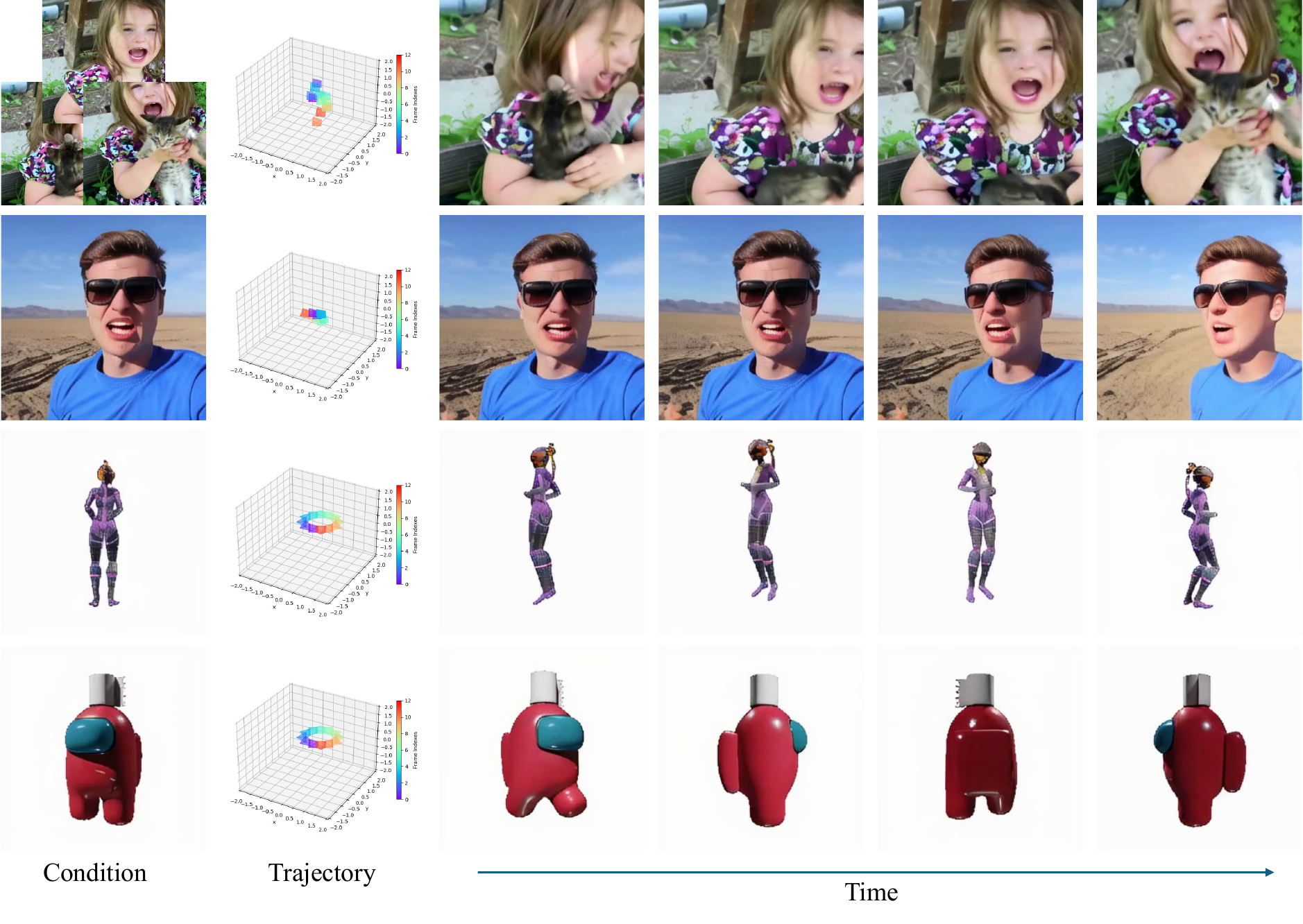}
    \caption{The visualization of the generated 4D videos. (Please refer to supplementary video for better illustration.)}
    \label{fig:4dgen-results}
\end{figure}

\begin{figure}[h]
    \centering
    \includegraphics[width=0.9\linewidth]{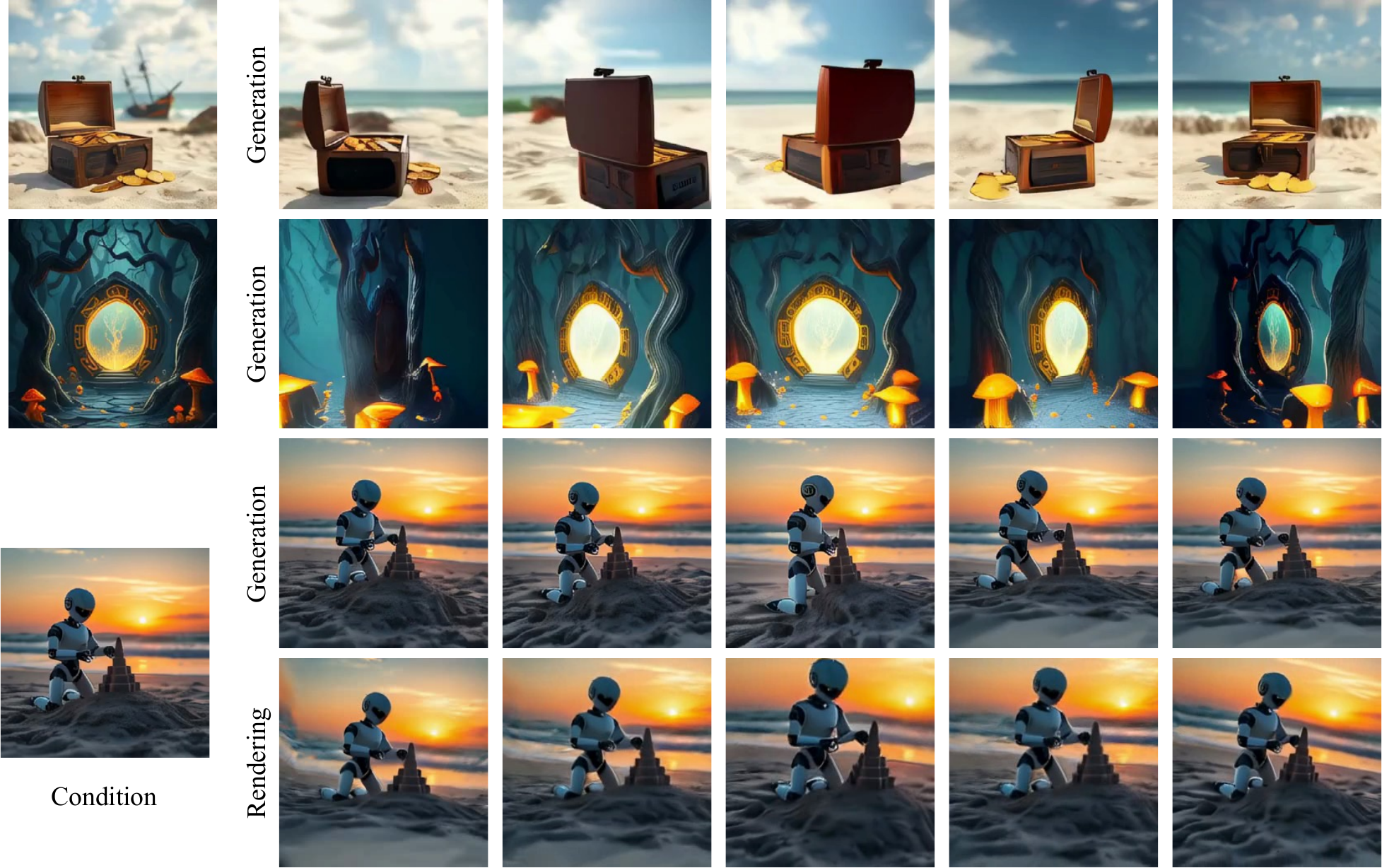}
    \caption{The visualization of the 3D generation results. The condition images are generated by FLUX.1. 
    The first three rows are the generated samples from \ours following camera trajectory, and the last row is the renderings from 3D-GS model trained with the generated samples. (Please refer to supplementary video for better illustration.)}
    \label{fig:3dgen-results}
\end{figure}

In this section, we show more qualitative results in general 3D and 4D generation with \ours.
In Fig.~\ref{fig:4dgen-results}, \ours can generate 3D and temporal consistent videos following the camera trajectory in with one or multiple condition images.
In addition, \ours can also achieve high quality 3D generation with in-the-wild images. In Fig.~\ref{fig:3dgen-results}, we use FLUX.1\footnote{https://huggingface.co/black-forest-labs/FLUX.1-schnell} to generate condition images and use \ours to generate multi-view images following orbit and forward-facing trajectories.
Furthermore, the generated samples are 3D consistent and we can lift them to 3D representations (\eg, 3D-GS). The last row of Fig.~\ref{fig:3dgen-results} is the renderings from 3D-GS model, which is optimized from the generated samples in the third row.

\begin{figure}
    \centering
    \includegraphics[width=0.9\linewidth]{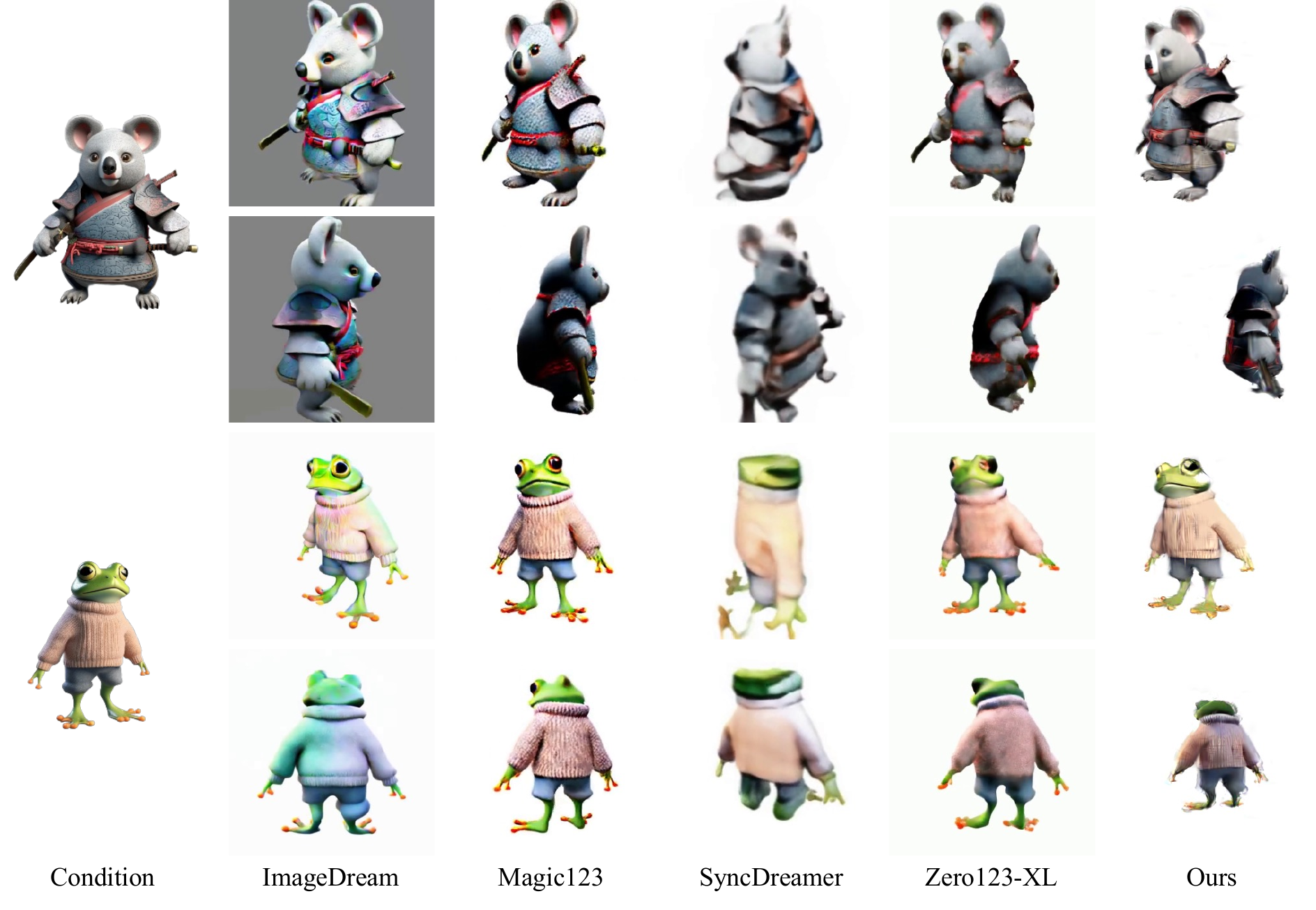}
    \caption{Qualitative comparison on single view 3D generation.}
    \label{fig:3dcomparison}
\end{figure}

\section{Single View 3D Generation}
\label{appendix:single-view}

\input{tables/single-view3d}

We evaluate \ours on the object-centric single view 3D generation. Specifically, we use the examples in \citet{wang2023imagedream} and calculate the evaluation metric with 7 views in the orbit trajectory.
The similarity between CLIP image embedding of the reference image and rendered views are adopted as evaluation metric.
%
As shown in Tab.~\ref{tab:single3d}, although our performance is slightly worse than IM-3D~\cite{im3d}, \ours achieves better performance than other baselines~\cite{one-2-3-45plusplus,wang2023imagedream} with a fast generation speed. In addition, \ours can handle scene-level generation and 4D generation, which cannot be achieved by other methods.
Furthermore, we only use basic 3D-GS model and train with the generated views. Approaches proposed by previous works~\cite{tang2023dreamgaussian,im3d} to improve the 3D quality can still be used. 
In Fig.~\ref{fig:3dcomparison}, we compare the rendering results from \ours with several previous methods~\cite{magic123,Zero-1-to-3,wang2023imagedream,liu2023syncdreamer}. The results of other methods are obtained from \citet{wang2023imagedream}. Compared with previous methods, \ours use the generated 3D consistent samples to optimize the 3D-GS model, instead of using distillation techniques~\cite{DreamFusion}. Therefore, \ours can well handle the over-saturated and Janus problem faced by other methods.

\section{Additional Ablation Studies}
\label{appendix:ablation}
In this section, we further analyze the effectiveness of ray camera condition and the joint training of 3D and 4D generation.

\textbf{Camera Condition.}
\ours leverages Pl\"ucker ray as camera condition, which is a dense pixel-wise camera representation. Some previous works~\cite{Zero-1-to-3,sargent2023zeronvs} convert the camera poses or spherical coordinate system to 1-dimension embeddings, and then integrate the condition into diffusion model with cross attention. As shown in the first row of Tab.~\ref{tab:ablation-appendix}, we convert the 3$\times$4 camera extrinsics and the focal length to 1-dimension embeddings and use it with cross attention layer (Camera CA). Compared to \ours, Camera CA performs worse on both 3D and 4D generation. In addition, due to the effectiveness of mask latent conditioned diffusion model, \ours do not requires cross attention in the denoising U-Net, which is more efficient.

\textbf{Joint training of 3D and 4D generation.}
\ours combine both 3D and 4D data into modeling training to facilitate 3D and 4D generation. In Tab.~\ref{appendix:ablation}, we conduct ablation studies on the effectiveness of each type of data. 3D data contains more camera position variation and more accurate camera poses. Thus, removing 3D data impairs both 3D and 4D generation, especially the generated 3D samples which cannot well aligned with the camera poses.
Instead, 4D data contains object motion together with spatial camera information. The lack of 4D data will impair the quality of the generated 3D samples. In addition, despite the performance drop on Cam-DAVIS is not significant in terms of FVD and FID, removing 4D data leads the model hardly generate object movement.

\input{tables/ablation-appendix}

\section{Details of Generation with 3D Representation}
\label{appendix:details}
For few-view 3D reconstruction, we fit trajectories from training views and generate samples corresponding to the cameras on the trajectory. Then the training views and generated samples are used together to optimize the 3D representation (3D Gaussian Splatting (3D-GS)~\cite{3dgs} and Zip-NeRF~\cite{barron2023zip}). 
Since both of them are designed for many views reconstruction, we modify the hyperparameter to fit for few-shot setting.
For Zip-NeRF, the width and depth of view-dependence network is set to 32 and 1 to avoid overfitting. The model is trained with batch size 262,144 (64$\times$64$\times$64) for 1k iterations. At each iteration, patch of 64$\times$64 are rendered and supervised with photometric and LPIPS loss.
For 3D-GS, the model is optimized for 10k iterations with photometric, SSIM and LPIPS loss. The initial point clouds are obtained from the input views for the baseline and all the generated views when combining it with \ours.
For single-view 3D generation, a set of anchor views are first generated with single view condition, and then we generate more views by interpolating the camera poses of these anchor views. All the generated views are directly used to optimize a 3D-GS model for 2k iterations.
4D generation requires the modeling of dynamic, so we use 4D Gaussian Splatting (4DGS)~\cite{4dgs} as backbone. Directly optimizing a 4D representation with 4D videos can be difficult to converge. Therefore, we first optimize a 3D-GS as 3D single-view generation manner and then optimize the dynamic deformation with the generated 4D video.

\section{Limitations}

\ours demonstrates remarkable performance in both 3D and 4D generation. However, there are two key limitations when applied to real-world scene generation. First, despite the abundance of 3D data, the diversity of real-world datasets is limited. For example, complex scenes~\cite{re10k}, typically feature simplistic camera trajectories, such as forward-facing views. In contrast, datasets focused on objects~\cite{mvimagenet,co3d}, often have more varied camera paths, providing richer 3D information, but the object categories are generally few in number and the scene structure is quite simple. As a result, \ours struggles with generating 360-degree views of complex scenes from single-view conditions.

Second, in 4D generation, temporal consistency and object motion are difficult to maintain during large camera movements in real-world scenarios. This limitation arises from the nature of available video data: video data typically provides limited object motion when the camera is moving quickly, whereas large object motion is often associated with static or slightly mobile cameras.
With such video data, our curated 4D data do not contain many samples with both large object and camera movement.

Both limitations are primarily due to the constraints of current datasets. However, with our proposed data curation pipeline and the increasing availability of public data, \ours has the potential for significant improvement in future applications.

\end{document}

%% file: math_commands.tex
\usepackage{amsmath,amsfonts,bm}

\def\eqref#1{equation~\ref{#1}}

\def\1{\bm{1}}

\DeclareMathAlphabet{\mathsfit}{\encodingdefault}{\sfdefault}{m}{sl}
\SetMathAlphabet{\mathsfit}{bold}{\encodingdefault}{\sfdefault}{bx}{n}



%% file: macros.tex
\definecolor{tabfirst}{rgb}{1, 0.7, 0.7}
\definecolor{tabsecond}{rgb}{1, 0.85, 0.7}
\definecolor{tabthird}{rgb}{1, 1, 0.7}

\newcommand{\maybeappendix}[1]{}

\newcommand\rgt{\aftergroup\mathclose\aftergroup{\aftergroup}\right}

\definecolor{brickred}{rgb}{0.8, 0.25, 0.33}
\definecolor{armygreen}{rgb}{0.29, 0.33, 0.13}
\definecolor{asparagus}{rgb}{0.53, 0.66, 0.42}
\definecolor{celadon}{rgb}{1.0, 0.64, 0.}
\newcommand{\cmark}{\textcolor{asparagus}{\ding{51}}}%
\newcommand{\xmark}{\textcolor{brickred}{\ding{55}}}%

\usepackage{array}
\newcolumntype{L}[1]{>{\raggedright\let\newline\\\arraybackslash\hspace{0pt}}m{#1}}
\newcolumntype{C}[1]{>{\centering\let\newline\\\arraybackslash\hspace{0pt}}m{#1}}
\newcolumntype{R}[1]{>{\raggedleft\let\newline\\\arraybackslash\hspace{0pt}}m{#1}}

\definecolor{antiquewhite}{rgb}{0.98, 0.92, 0.84}

\newcommand{\ours}{Gen$\mathcal{X}$D\xspace}
\newcommand{\data}{CamVid-30K\xspace}

\def\eg{\emph{e.g.}}

%% file: tables/teaser-table.tex
\begin{table}[t]
    \centering
    \caption{Comparison among the settings of previous works.}
    \label{tab:setting}
    \vspace{-.1in}
    \resizebox{.9\textwidth}{!}{
    \begin{tabular}{l|cccc|cccc}
        \toprule
        \multirow{2}{*}{\textbf{Method}} & \multicolumn{4}{c|}{\textbf{3D Generation}} & \multicolumn{4}{c}{\textbf{4D Generation}} \\
        & \textbf{Object} & \textbf{Scene} &\textbf{Single View} & \textbf{Multi-View} & \textbf{Object} & \textbf{Scene} &\textbf{Single View} & \textbf{Multi-View} \\
        \midrule
        IM-3D & \cmark & \xmark & \cmark & \xmark & \xmark & \xmark & \xmark & \xmark \\
        RealmDreamer & \xmark & \cmark & \cmark & \xmark & \xmark & \xmark & \xmark & \xmark \\
        ReconFusion & \cmark & \cmark & \xmark & \cmark & \xmark & \xmark & \xmark & \xmark \\
        CAT3D &  \cmark & \cmark & \cmark & \cmark & \xmark & \xmark & \xmark & \xmark \\
        Animate124 & \xmark & \xmark & \xmark & \xmark & \cmark & \xmark & \cmark & \xmark  \\
        CameraCtrl & \xmark & \xmark & \xmark & \xmark & \xmark & \cmark & \cmark & \xmark  \\
        SV4D & \cmark & \xmark & \cmark & \xmark & \cmark & \xmark & \cmark & \cmark \\
        CamCo & \xmark & \cmark & \cmark & \xmark & \xmark & \cmark & \cmark & \xmark \\
        \midrule
        \textbf{\ours} (Ours) & \cmark & \cmark & \cmark & \cmark & \cmark & \cmark & \cmark & \cmark \\
        \bottomrule
    \end{tabular}}
    \vspace{-.15in}
\end{table}

%% file: tables/4d.tex
\begin{table}[t]
\centering
\small

\begin{minipage}{0.45\textwidth}
    \centering
    \caption{\textbf{4D scene generation.}}
    \vspace{-.1in}
    \label{tab:4d-scene}
    \resizebox{0.99\textwidth}{!}{%
    \begin{tabular}{l|ccc}
    \toprule
    Method & FID $\downarrow$  & FVD $\downarrow$\\
    \midrule
    MotionCtrl~\cite{wang2024motionctrl} & 118.14 & 1464.08 \\
    CameraCtrl~\cite{he2024cameractrl} & 138.64 & 1470.59 \\
    \midrule
    \ours (Single View) & 101.78 & 1208.93 \\
    \ours (3 Views) & \textbf{55.64} & \textbf{490.50} \\
    \bottomrule
    \end{tabular}}
    
\end{minipage}
\hfill
\begin{minipage}{0.54\textwidth}
    \centering
    \caption{\textbf{4D object generation.}}
    \vspace{-.1in}
    \label{tab:4d-object}
    \resizebox{0.99\textwidth}{!}{%
    \begin{tabular}{l|cc}
    \toprule
    Method & Time $\downarrow$ & CLIP-I $\uparrow$  \\
    \midrule
    Zero-1-to-3-V~\cite{Zero-1-to-3} & 4 hrs & 79.25 \\
    RealFusion-V~\cite{RealFusion} & 5 hrs & 80.26 \\
    Animate124~\cite{zhao2023animate124} & 7 hrs & 85.44\\
    \midrule
    \ours (Single View) & \textbf{4 min} & \textbf{90.32} \\
    \bottomrule
    \end{tabular}}
    
\end{minipage}
\label{tab:4d}
\vspace{-.1in}
\end{table}

%% file: tables/few-view3d.tex
\begin{table}[t]
    \caption{\textbf{Quantitative comparison of few-view 3D reconstruction} on both in-distribution (Re10K) and out-of-distribution (LLFF) datasets.}
    \vspace{-.1in}
    \small
    \label{tab:few-view3d}
    \centering
    \resizebox{.8\textwidth}{!}{
    \begin{tabular}{l|ccc|ccc}
    \toprule
    \multirow{2}{*}{\textbf{Method}} & \multicolumn{3}{c|}{\textbf{Re10K}} & \multicolumn{3}{c}{\textbf{LLFF}} \\
    & PSNR$\uparrow$ & SSIM$\uparrow$ & LPIPS$\downarrow$ & PSNR$\uparrow$ & SSIM$\uparrow$ & LPIPS$\downarrow$ \\
    \midrule
    Zip-NeRF~\cite{barron2023zip}  & 20.58 & 0.729 & 0.382 & 14.26 & 0.327 & 0.613 \\
    \textbf{Zip-NeRF + {\ours}} & \textbf{25.40} & \textbf{0.858} & \textbf{0.223 }& \textbf{19.39} & \textbf{0.556} & \textbf{0.423} \\
    \midrule
    3D-GS~\cite{3dgs}   & 18.84 & 0.714 & 0.286 & 17.35 & 0.489 & 0.335 \\
    \textbf{3D-GS + {\ours}} & \textbf{23.13} & \textbf{0.808} & \textbf{0.202} & \textbf{19.43} & \textbf{0.554} & \textbf{0.312} \\
    \bottomrule
    \end{tabular}}
\vspace{-.2in}
\end{table}

%% file: tables/ablation.tex
\begin{table}[t]
\small
    \caption{\textbf{Ablation studies on motion disentangle.}}
    \vspace{-.1in}
    \label{tab:ablation}
    \centering
    \resizebox{.9\textwidth}{!}{
    \begin{tabular}{l|ccc|ccc|cc}
    \toprule
    \multirow{2}{*}{\textbf{Method}} & \multicolumn{3}{c|}{\textbf{Re10K}} & \multicolumn{3}{c|}{\textbf{LLFF}} & \multicolumn{2}{c}{\textbf{Cam-DAVIS}} \\
    & PSNR$\uparrow$ & SSIM$\uparrow$ & LPIPS$\downarrow$ & PSNR$\uparrow$ & SSIM$\uparrow$ & LPIPS$\downarrow$ & FID$\downarrow$ & FVD$\downarrow$ \\
    \midrule
    w.o. Motion Disentangle & 20.75 & 0.635 & 0.362 & 16.89 & 0.397 & 0.560 & 122.73 & 1488.47 \\
    \textbf{\ours} & \textbf{22.96} & \textbf{0.774} & \textbf{0.341} & \textbf{17.94} & \textbf{0.463} & \textbf{0.546} & \textbf{101.78}  & \textbf{1208.93} \\
    \bottomrule
    \end{tabular}}
\vspace{-.15in}
\end{table}

%% file: tables/single-view3d.tex
\begin{table}[t]
\centering
\caption{Quantitative comparison of image-to-3D generation on examples from \citet{wang2023imagedream}.}
\vspace{-.1in}
\small
    \begin{tabular}{l|c|cc}
    \toprule
    Method & Model Type &  Time (min) $\downarrow$ & CLIP-I $\uparrow$ \\
    \midrule
    ImageDream~\cite{wang2023imagedream} & 3D & 120 & 83.77 \\
    One2345++~\cite{one-2-3-45plusplus} & 3D & \textbf{0.75} & 83.78  \\
    IM-3D~\cite{im3d} & 3D & 3 & \textbf{91.40 } \\
    \midrule
    \ours (ours) & 3D\&4D & 2 & 84.75 \\
    \bottomrule
    \end{tabular}
\label{tab:single3d}
\end{table}

%% file: tables/ablation-appendix.tex
\begin{table}[t]
\small
    \caption{Ablation studies on camera conditioning scheme and joint training. }
    \vspace{-.1in}
    \label{tab:ablation-appendix}
    \centering
    \resizebox{.99\textwidth}{!}{
    \begin{tabular}{l|ccc|ccc|cc}
    \toprule
    \multirow{2}{*}{\textbf{Method}} & \multicolumn{3}{c|}{\textbf{Re10K}} & \multicolumn{3}{c|}{\textbf{LLFF}} & \multicolumn{2}{c}{\textbf{Cam-DAVIS}} \\
    & PSNR$\uparrow$ & SSIM$\uparrow$ & LPIPS$\downarrow$ & PSNR$\uparrow$ & SSIM$\uparrow$ & LPIPS$\downarrow$ & FID$\downarrow$ & FVD$\downarrow$ \\
    \midrule
    Camera CA & 21.73  & 0.692 & 0.355 & 17.15 & 0.434 & 0.573 & 105.69 & 1331.62 \\
    w.o. 3D Data & 16.38 & 0.604 & 0.465 & 14.98 & 0.400 & 0.587 & 107.74 & 1262.12 \\
    w.o. 4D Data & 20.74 & 0.740 & 0.359 & 17.35 & 0.448 & 0.554 & 107.93 & 1240.57 \\
    \textbf{\ours} & \textbf{22.96} & \textbf{0.774} & \textbf{0.341} & \textbf{17.94} & \textbf{0.463} & \textbf{0.546} & \textbf{101.78}  & \textbf{1208.93} \\
    \bottomrule
    \end{tabular}}
\vspace{-.1in}
\end{table}